\documentclass[conference]{IEEEtran}
\IEEEoverridecommandlockouts

\usepackage{cite}
\usepackage{amsmath,amssymb,amsfonts}
\usepackage{algorithmic}
\usepackage{graphicx}
\usepackage{textcomp}
\usepackage{xcolor}
\usepackage{booktabs}
\usepackage{multirow}
\usepackage{array}
\usepackage{tabularx}
\usepackage{makecell}
\usepackage{placeins}
\usepackage{dblfloatfix}
\usepackage{url}
\newcolumntype{Y}{>{\raggedright\arraybackslash}X}
\newcolumntype{L}[1]{>{\raggedright\arraybackslash}p{#1}}

\def\BibTeX{{\rm B\kern-.05em{\sc i\kern-.025em b}\kern-.08em
    T\kern-.1667em\lower.7ex\hbox{E}\kern-.125emX}}

\begin{document}

\title{One-Stage Object Detectors in Autonomous Driving}

\author{
\IEEEauthorblockN{Jonel Roman, Ryan Sirjue, Peter Nguyen, Daniel Krutky, Juan Jesus, Sudip Dhakal}
\IEEEauthorblockA{
CEN 4930 - CRN 16785 \\
Florida Gulf Coast University 
}
}

\maketitle

\begin{abstract}
\textbf{Autonomous vehicles depend on fast and reliable perception systems to detect surrounding vehicles, pedestrians, cyclists, traffic signs, and other road objects in real time. This paper presents a comprehensive survey and analysis of one-stage object detectors for autonomous driving rather than an implementation of a new detection system. The survey reviews the evolution of major one-stage detectors, including YOLOv1, SSD, RetinaNet, EfficientDet, anchor-free detectors such as FCOS and CenterNet, and recent real-time models such as YOLOv10. It compares these architectures through their design choices, feature-fusion strategies, loss functions, deployment trade-offs, and reported benchmark performance. The paper also summarizes commonly used autonomous-driving datasets, evaluation metrics, open challenges, and future research directions. Overall, this survey highlights how one-stage detectors balance speed, accuracy, efficiency, and robustness, while also emphasizing the remaining gap between benchmark results and dependable real-world autonomous-driving performance.}
\end{abstract}

\begin{IEEEkeywords}
Autonomous driving, one-stage object detection, perception, real-time detection, survey
\end{IEEEkeywords}

\section{Introduction}

Object detection is a core component of the perception module in autonomous vehicles. A driving system must identify surrounding vehicles, pedestrians, cyclists, traffic signs, lane obstacles, and other road users quickly enough to support tracking, planning, and control decisions. In this context, detection results are not only visual outputs; bounding boxes, class labels, and confidence scores influence how the vehicle estimates risk and responds to its environment. Therefore, autonomous-driving detection must balance accuracy with real-time latency constraints, especially in fast-changing road scenes \cite{sarda2021yolo,liu2024largeai}. 

Deep learning-based object detection is commonly grouped into two major paradigms: two-stage detection and one-stage detection. Two-stage detectors, such as the R-CNN family, first generate candidate regions and then classify and refine those regions. This design can produce strong localization accuracy, but the region-proposal step increases computational cost and can limit real-time deployment. One-stage detectors remove the separate proposal stage and directly predict bounding boxes and class probabilities in a single forward pass, making them attractive for autonomous-driving perception where low latency is essential \cite{redmon2016yolo, liu2016ssd, lin2017focalloss}.

This paper is a survey and comparative analysis, not a report on a newly implemented detection system. Its purpose is to organize the development of one-stage object detectors, compare their architectural choices and reported performance, and discuss how their strengths and weaknesses relate to autonomous-driving use cases. Recent models such as EfficientDet and YOLOv10 show that the field is moving beyond simple speed improvements toward better feature fusion, reduced post-processing overhead, and more efficient deployment. However, real-world autonomous driving still requires stronger robustness to small objects, occlusion, lighting variation, weather, and hardware limitations \cite{tan2020efficientdet, wang2024yolov10, zeng2020refpn, lee2022badweather}.

\begin{table*}[!t]
\caption{Taxonomy of Object Detection Methods in Autonomous Driving}
\label{tab:taxonomy}
\centering
\scriptsize
\renewcommand{\arraystretch}{1.2}
\setlength{\tabcolsep}{4pt}

\begin{tabularx}{\textwidth}{L{0.18\textwidth} L{0.18\textwidth} L{0.18\textwidth} Y}
\toprule
\textbf{Paradigm} & \textbf{Architectural Family} & \textbf{Sub-Family Classification} & \textbf{Representative Architectures} \\
\midrule

Two-Stage Detectors
& R-CNN Series
& Region Proposal
& Faster R-CNN, Mask R-CNN, Cascade R-CNN  \\[0.8em]

\multirow{4}{=}{One-Stage Detectors}
& \multirow{2}{=}{Anchor-Based}
& Default Box Matching
& SSD, RetinaNet, EfficientDet \cite{liu2016ssd, lin2017focalloss, tan2020efficientdet} \\[0.8em]

& 
& Grid-Based Anchors
& YOLOv2, YOLOv3, YOLOv4, YOLOv5, YOLOv6, YOLOv7 \cite{redmon2017yolo9000, redmon2018yolov3, bochkovskiy2020yolov4, wang2023yolov7} \\[0.8em]

& \multirow{2}{=}{Anchor-Free}
& Center-Based
& YOLOv1, FCOS, YOLOX, YOLOv8, YOLOv9, YOLOv10 \cite{redmon2016yolo, tian2019fcos, ge2021yolox, wang2024yolov10} \\[0.8em]

&
& Keypoint-Based
& CornerNet, CenterNet, ExtremeNet \cite{law2018cornernet, zhou2019objects} \\

\bottomrule
\end{tabularx}
\end{table*}

This survey provides a comprehensive synthesis of the field, defined by the following specific contributions:

\begin{enumerate}
    \item To provide a chronological order and analysis of foundational one-stage detectors that detail innovations.
    \item We evaluate and benchmark the speed-accuracy trade-offs of these models on specific autonomous driving datasets, including KITTI, Waymo, and BDD100K.
    \item We detail the practical advantages, disadvantages, and specific deployment limitations of each detector regarding resource-constrained edge hardware.
    \item We identify open challenges in the field, such as small object detection and adverse weather robustness, and propose concrete future research directions to bridge the gap between benchmark accuracy and real-world autonomous vehicle dependability.
\end{enumerate}

\section{Literature Review}

This section combines prior work on one-stage object detectors in autonomous driving by sorting the literature around major themes that appear across the selected papers. Rather than reviewing each paper separately, this section connects the studies through their shared design goals, trade-offs, and unresolved challenges. \cite{arnold2021onperformance,balasubramaniam2022status}.

\subsection{Motivation for One-Stage Detection in Autonomous Driving}

The role of one-stage detection in autonomous driving is increasingly defined by its ability to deliver fast, accurate perception under the strict latency demands of real-world driving. Across the selected papers, this role is developed from several complementary angles. EfficientDet shows that one-stage detectors can remain highly competitive by improving feature fusion and scaling efficiency, making strong performance possible even under limited computational budgets. ReFPN-FCOS contributes by strengthening localization quality, reinforcing the idea that autonomous-driving perception depends not only on detecting objects quickly, but also on placing them precisely in space. YOLOv10 extends the efficiency argument further by targeting bottlenecks such as non-maximum suppression, suggesting that the future role of one-stage detectors lies in reducing procedural overhead as much as improving backbone performance. Finally, the adverse-weather paper places these architectural gains into the autonomous-driving context, showing that one-stage detectors are valuable only if they remain dependable in dynamic and degraded driving scenes. Across these works, one-stage detection emerges as a practical foundation for autonomous-driving perception because the models increasingly combine low latency with improved localization and robustness \cite{tan2020efficientdet,zeng2020refpn,wang2024yolov10,lee2022badweather}.

\subsection{Real-Time Efficiency and Speed--Accuracy Trade-Offs}
Research on real-time efficiency has moved from broad detector comparisons toward more application-specific studies, including pothole detection, motorbike detection, and autonomous-driving benchmarks. Across these works, YOLO-style models generally show the strongest potential for real-time deployment because they maintain high throughput while preserving usable detection accuracy. However, the studies also show that speed alone is not enough: an autonomous vehicle needs detectors that remain reliable across diverse road scenes, object scales, and environmental conditions \cite{yilmaz2025realtime,yinkfu2025motorbike,pothole2025yolov5ssd,sarda2021yolo}. 

\subsection{Architectural and Methodological Innovations}
Research on architectural innovations in one-stage detectors has progressed from earlier anchor-based designs such as YOLOv3 and SSD, which emphasized real-time speed, toward more refined architectures that balance efficiency and detection quality through deliberate changes to the backbone, neck, head, and feature-fusion pipeline. Across the selected papers, a consistent pattern is the effort to improve speed without giving up localization accuracy, whether through revised loss and assignment strategies, stronger feature selection, decoupled prediction heads, anchor-free design, or re-parameterized and efficiency-oriented modules. This line of work shows that progress in one-stage detection depends on redesigning core components so high-speed inference can remain compatible with the precision and robustness required in autonomous-driving systems \cite{redmon2018yolov3,liu2016ssd,ge2021yolox,feng2021tood,lyu2022rtmdet}.

\subsection{Detection Accuracy in Autonomous Driving Environments}
Research on detection accuracy, alignment, and optimization strategy has evolved from earlier speed-centered one-stage methods (like early YOLO versions) toward more refined approaches that improve prediction quality through targeted methodological changes. Across the selected papers, a persistent pattern is the attempt to strengthen real-time detection not only through faster inference, but through better loss design, more effective label assignment, and improved alignment between classification confidence and localization precision. Rather than relying only on larger or faster architecture, these studies show that detection quality can be significantly improved through deliberate optimization choices, including IoU-aware supervision and Distribution Focal Loss that better distinguish high-quality matches. For autonomous driving, these optimization-focused studies are important because they improve the reliability of detections without relying only on larger or slower architectures \cite{zhao2018dynamic,li2020gfl,zhang2021vfnet,feng2021tood}.

\subsection{AV-Specific Challenges and Practical Limitations}

Research on this subtopic is wide and varied because autonomous vehicles still face many challenges in real-world driving environments. Across these papers, different solutions are proposed for different problems, such as improving detection accuracy, handling difficult road conditions, and making perception systems more reliable. These works show that real-world AV perception is a system-level challenge: detector accuracy must be supported by robust sensing, diverse training data, and careful deployment choices rather than by a single model improvement alone \cite{balasubramaniam2022status,du2023object,lee2022badweather,liu2023lidarcamera}.

\subsection{Emerging Trends and Research Gaps}
Across some of the selected papers, an important emerging trend is the shift from simply making one-stage detectors faster toward making them more accessible through improvements in alignment, efficiency, scaling, and real-time optimization. TOOD reflects this trend by addressing the misalignment between classification confidence and localization precision, while EfficientDet emphasizes scalable efficiency through BiFPN and compound scaling, and RTMDet pushes further toward industrial-grade real-time performance through architectural refinement, improved label assignment, and optimized training strategies. At the same time, the Waymo-based comparative study shows that these advances do not fully remove the practical difficulties of autonomous-driving perception, since real deployment still requires robustness to occlusion, class imbalance, adverse conditions, and strict latency constraints. The newer methods reduce several older weaknesses, but they often shift the trade-offs rather than eliminating them completely. Benchmark accuracy and efficiency continue to improve, yet the gap between strong general-purpose results and dependable real-world AV performance remains open. This gap motivates a survey perspective that connects architectural progress with deployment-oriented limitations \cite{feng2021tood,tan2020efficientdet,lyu2022rtmdet,arnold2021onperformance}.

\section{Background and Technical Taxonomy}

\subsection{Core Concepts to Define}
\textbf{Feature Fusion (BiFPN)}: As demonstrated by EfficientDet, this involves bidirectional cross-scale connections and weighted fusion to better integrate multi-scale features for complex object detection \cite{tan2020efficientdet}.

\textbf{Localization Accuracy vs. Confidence}: A critical distinction highlighted in the ReFPN-FCOS research; high classification confidence doesn't always guarantee spatial precision, which is vital for safety-critical driving tasks \cite{zeng2020refpn}.

\textbf{Class Imbalance and Training Convergence}: The challenge where background data outweighs object data. Research into Dynamic Loss (DL) functions seeks to solve this by focusing on learning "hard" examples \cite{lin2017focalloss,zhao2018dynamic}. 

\textbf{End-to-End Detection}: A paradigm seen in YOLOv10 that eliminates post-processing bottlenecks like Non-Maximum Suppression (NMS) to achieve real-time inference \cite{wang2024yolov10}

\subsection{Three Sub-Families of One-Stage Detectors}

\begin{table}[htbp]
\caption{Three sub-families of one-stage detectors}
\label{tab:subfamilies}
\centering
\small
\renewcommand{\arraystretch}{1.25}
\begin{tabularx}{\columnwidth}{L{0.26\columnwidth} Y Y}
\toprule
\textbf{Sub-Family} & \textbf{Core Design Principle} & \textbf{Key Detectors That Belong Here} \\
\midrule
Anchor-Free / Point-Based & Eliminates predefined anchor boxes to reduce complexity and improve localization of irregular objects & FCOS (ReFPN-FCOS) \\
Efficiency-Optimized & Focuses on reducing parameter counts and FLOPs for deployment on edge hardware. & EfficientDet, ShuffYOLOX \\
NMS-Free / Real-Time YOLO & Utilizes dual-assignment strategies to remove inference bottlenecks and reach ultra-low latency. & YOLOv10, YOLOX \\
\bottomrule
\end{tabularx}
\end{table}

\section{Overview of One-Stage Detectors}

\begin{figure*}[!t]
\centering
\includegraphics[width=0.98\textwidth]{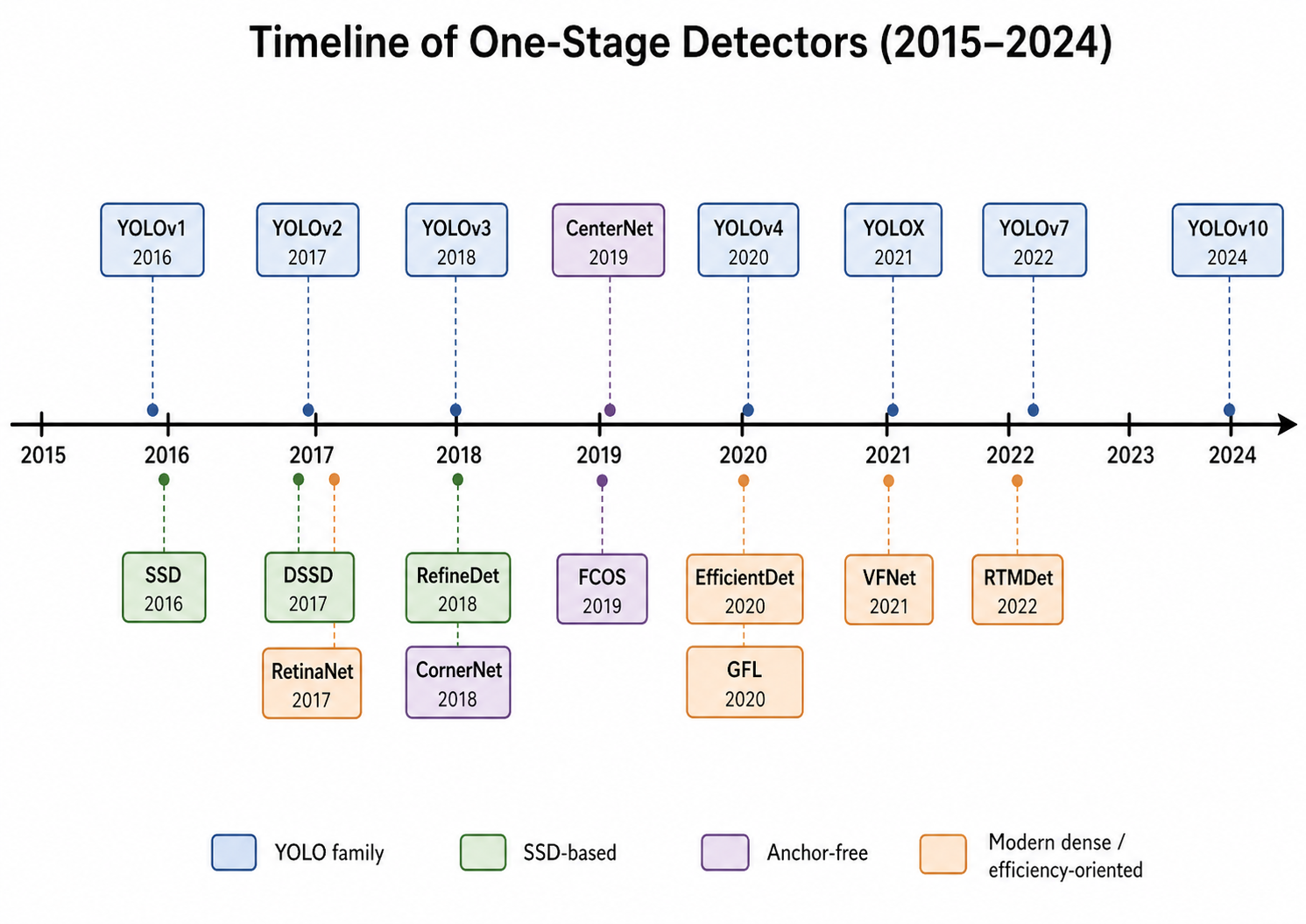}
\caption{Timeline of one-stage detectors from 2016 to 2024. The figure is placed across both columns to improve readability. If the source graphic is regenerated, use larger labels and more spacing between model boxes.}
\label{fig:timeline}
\end{figure*}

Fig.~\ref{fig:timeline} shows the chronological evolution of the selected one-stage detectors.

\begin{table*}[!t]
\caption{Summary of the selected one-stage detectors}
\label{tab:detector_summary}
\centering
\scriptsize
\renewcommand{\arraystretch}{1.18}
\setlength{\tabcolsep}{4pt}
\begin{tabularx}{\textwidth}{c L{0.10\textwidth} c L{0.20\textwidth} Y}
\toprule
\textbf{\#} & \textbf{Detector} & \textbf{Year} & \textbf{Sub-Family} & \textbf{Key Innovation} \\
\midrule
1  & YOLOv1    & 2016 & YOLO Family & Increasing the real-time speed and detection accuracy \\
2  & YOLOv2    & 2017 & YOLO Family & Better accuracy \\
3  & YOLOv3    & 2018 & YOLO Family & Multi-scale prediction \\
4  & YOLOv4    & 2020 & YOLO Family & An increase in detection accuracy \\
5  & YOLOX     & 2021 & YOLO Family & An anchor-free formulation and SimOTA label assignment \\
6  & YOLOv7    & 2023 & YOLO Family & Architectural refinements \\
7  & YOLOv10   & 2024 & YOLO Family & Improved efficiency and accuracy \\
\addlinespace[2pt]
8  & SSD       & 2016 & SSD-Based Detectors & Predict category scores and box offsets \\
9  & DSSD      & 2017 & SSD-Based Detectors & Improving the strength of detecting smaller objects \\
10 & RefineDet & 2018 & SSD-Based Detectors & two-step refinement strategy in a one-stage detector \\
\addlinespace[2pt]
11 & CornerNet & 2018 & Anchor-Free One-Stage Detectors & Defined key points for the object instead of the use of bounding-boxes \\
12 & CenterNet & 2019 & Anchor-Free One-Stage Detectors & Creating center points for anchor-free detection \\
13 & FCOS      & 2019 & Anchor-Free One-Stage Detectors & Fully convolutional anchor-free detection with per-pixel box regression and centerness scoring \\
\addlinespace[2pt]
14 & RetinaNet    & 2017 & Modern Dense and Efficiency-Oriented Detectors & Focal loss from the foreground-background imbalances \\
15 & EfficientDet & 2020 & Modern Dense and Efficiency-Oriented Detectors & BiFPN \\
16 & GFL          & 2020 & Modern Dense and Efficiency-Oriented Detectors & Clearing the inconsistencies in classification and localization \\
17 & VFNet        & 2021 & Modern Dense and Efficiency-Oriented Detectors & IoU-Aware classification score and Varifocal Loss \\
18 & RTMDet       & 2022 & Modern Dense and Efficiency-Oriented Detectors & A more balanced backbone-neck design \\
\bottomrule
\end{tabularx}
\end{table*}

Following the summary provided in Table~\ref{tab:detector_summary}, the selected one-stage detectors are discussed in greater detail. To maintain a clear and systematic organization, the detectors are grouped by sub-family and arranged chronologically within each group. For each detector, the discussion focuses on its original publication, key innovation, architectural design, improvements over prior methods, remaining limitations, and relevance to autonomous driving.

\subsection{YOLO Family}
The YOLO family is one of the most influential branches of one-stage object detection. Across its successive versions, these detectors have aimed to improve the balance between real-time speed and detection accuracy, making them especially important for autonomous driving applications \cite{redmon2016yolo,redmon2017yolo9000,redmon2018yolov3,bochkovskiy2020yolov4,ge2021yolox,wang2023yolov7,wang2024yolov10}.

\subsubsection{YOLOv1}
YOLOv1 was introduced in 2015 at CVPR and belongs to the YOLO family of one-stage detectors. Its main contribution was to frame detection as a single regression problem from image pixels to bounding boxes and class probabilities, rather than using region proposals. It showed that real-time detection was possible.

Architecturally, YOLOv1 divides the image into regions and predicts boxes, confidence scores, and class probabilities in one forward pass. Its loss handles localization, confidence, and classification, but the model struggles with smaller objects, multiple objects in frame, and precise localization. It is important because it established the real-time detection idea \cite{redmon2016yolo}.

\subsubsection{YOLOv2}
YOLOv2 was introduced in 2017 in YOLO9000. Its core contribution was improving YOLOv1’s accuracy while preserving the speed through anchor boxes, batch normalization, dimension clusters, high-resolution classification, and multi-scale training. When compared with YOLOv1, it was made to reduce the weakness of localization and improve the detection.

The architecture for YOLOv2, uses Darknet-19 as its backbone and predicts anchor-based detections at its final stage. The training objective still combines classification, localization, and confidence terms, while multi-scale training improves the adaptability of the input sizes. Though the speed-accuracy has increased, it still has difficulty with small objects and complex multi-scale scenes \cite{redmon2017yolo9000}.

\subsubsection{YOLOv3}
YOLOv3 was introduced in 2018 as an improvement to the YOLO line. Its main innovation was the multi-scale prediction together with a stronger backbone, Darknet-53, which helped the detector handle different sized objects more effectively. Compared with YOLOv2, it has moved the series closer to a modern dense detector because of the improved accuracy,

Architecturally, YOLOv3 uses residual connections in Darknet-53 and predicts detections at three scales, which is useful for small objects from afar. The model separates objectness, class prediction, and bounding-box regression, but it still trails the stronger detectors in localization and performance in heavily occluded scenes \cite{redmon2018yolov3}.

\subsubsection{YOLOv4}
YOLOv4 was introduced in 2020 and focused on combining many practical training and architectural improvements into a strong detector. Its main contribution was not one single module, it was a collection of modules such as CSP connections, Mosaic augmentation, Mish activation, and CIoU-based optimization. These modules helped push detection accuracy much higher without requiring certain hardware.

Architecturally, YOLOv4 uses CSPDarknet53 as the backbone, PANet-style neck components, and a YOLO detection head. Its training losses and optimization strategies intended to improve localization and generalization, under certain conditions. Even so, it remains anchor-based and can still struggle in highly crowded scenes and small-object cases \cite{bochkovskiy2020yolov4}.

\subsubsection{YOLOX}
YOLOX was introduced in 2021 as a modernized YOLO-style detector. Its main innovations were switching to an anchor-free formulation, using a decoupled head, and adopting SimOTA label assignment. These changes were intended to modernize the YOLO family and close the gap with stronger contemporary dense detectors while keeping practical speed. 

Architecturally, YOLOX keeps a YOLO-like real-time pipeline but separates classification and regression more explicitly in the head and abandons anchors. Its training setup is a major reason for its performance gains. Although it is strong and practical, it still inherits the broader challenge of balancing latency, small-object sensitivity, and robustness under difficult weather or occlusion. In autonomous driving, YOLOX is especially appealing because it was successful in streaming perception contexts and fits real-time constraints well \cite{ge2021yolox}.

\subsubsection{YOLOv7}
YOLOv7 was introduced in 2022 and later appeared in CVPR 2023. Its central contribution was the idea of trainable set of tools together with architectural refinements that improved both speed and accuracy in the real-time system. It aimed to outperform existing real-time detectors across a broad FPS range without abandoning the practicality expected from the YOLO family.

Architecturally, YOLOv7 refines the backbone, neck, and training strategy rather than replacing the whole pipeline with a new formulation. The emphasis is on optimization strategies that raise performance while keeping inference efficient. Like other real-time detectors, it can still face challenges with severe occlusion, domain shift, and very small objects. For autonomous driving, YOLOv7 is important because it represents one of the strongest high-speed detectors of its generation \cite{wang2023yolov7}.
\subsubsection{YOLOv10}
YOLOv10 was introduced in 2024 as a modern member of the YOLO family designed for real-time object detection. Its main contribution was improving efficiency and accuracy while moving toward a non-maximum-suppression-free detection pipeline using consistent dual assignments during training. Compared with earlier YOLO variants, YOLOv10 aimed to reduce post-processing overhead and strengthen the speed-accuracy tradeoff, making it an important recent step in the evolution of one-stage detectors for applications.

Architecturally, YOLOv10 refines multiple components of the detector to improve both computational efficiency and detection performance rather than depending on a single major design change. The model emphasizes end-to-end detection, reduced latency, and strong efficiency across different model scales. Although YOLOv10 represents a strong modern detector, it still must be considered in the context of difficult autonomous driving conditions such as occlusion, small distant objects, and adverse weather. In autonomous driving, its importance comes from its focus on low-latency and high-efficiency perception, both of which are critical for safe operation in dynamic road environments \cite{wang2024yolov10}.

\subsection{SSD-Based Detectors}
SSD-based detectors represent another major branch of one-stage detection. These models emphasized dense multi-scale prediction and helped establish the practical foundation for later real-time detectors used in complex visual environments such as road scenes \cite{liu2016ssd,fu2017dssd,zhang2018refinedet}.

\subsubsection{SSD}
SSD was introduced in 2016 as Single Shot MultiBox Detector. Its main innovation is to predict the category scores and box offsets from multiple feature maps at different scales using default boxes, allowing the one-stage detection while improving the single-shot attempts.

Architecturally, SSD uses a base network with an extra convolutional layer to generate predictions at different resolutions. Its loss combines the classification and localization terms over matched default boxes. Though SSD is fast, it is weaker on small-object detection \cite{liu2016ssd}.

\subsubsection{DSSD}
DSSD was introduced in 2017 as an extension of SSD. Its main innovation was adding deconvolution-based modules and extra context to improve the original SSD’s weakness on small objects and the understanding of richer scenes.

Architecturally, DSSD combines an SSD-style framework with a stronger classifier backbone and deconvolution layers for context aggregation. Its detection objective remains a multi-task combination of classification and localization losses. The tradeoff is that DSSD has improved accuracy for small objects, but the complexity is increased at the sacrifice of speed \cite{fu2017dssd}.

\subsubsection{RefineDet}
RefineDet was introduced in 2018 to narrow the gap between the efficiency of one-stage and two-stage detectors. Its core contribution was the two-step refinement strategy within a one-stage pipeline. It allowed for an anchor refinement Module that adjusts the anchors, then and object detection module that performs the final classification and regression. It was done to reduce the search space and improve localization.

Architecturally, RefineDet includes anchor refinement, transfer connection blocks, and a final detection stage trained end-to-end with multi-task loss. Compared with SSD-style detectors, it improves accuracy, but it is more complex and less conceptually simple than earlier predictors. RefineDet is crucial because it improves localization quality for road users \cite{zhang2018refinedet}.

\subsection{Anchor-Free One-Stage Detectors}
Anchor-free one-stage detectors were developed to reduce the complexity of anchor-box design and to provide more flexible detection strategies. These methods are particularly relevant in autonomous driving because road objects vary significantly in size, shape, and spatial arrangement \cite{law2018cornernet,zhou2019objects,tian2019fcos}.

\subsubsection{CornerNet}
CornerNet was introduced in 2018 as an anchor-free one-stage object detector that represented objects as pairs of key points rather than relying on predefined anchor boxes. Its main innovation was to detect the top-left and bottom-right corners of each object and then group them together to form bounding boxes. This was important because it offered a different way of thinking about object detection, avoiding the anchor settings used in many earlier one-stage models. Compared with anchor-based detectors, CornerNet aimed to simplify the detection formulation while still achieving strong accuracy, making it an influential step in the development of anchor-free detection methods.

Architecturally, CornerNet predicts corner heatmaps, embedding vectors used to pair matching corners, and offset values to improve localization precision. It also introduced corner pooling to better localize object boundaries by gathering information along horizontal and vertical directions. Although CornerNet was innovative and influential, it can be harder to deploy efficiently in crowded scenes where many corners must be correctly matched, and later anchor-free methods often provided simpler or faster alternatives. In autonomous driving, CornerNet is important because it helped push the field away from strict reliance on anchor boxes and inspired later detectors that are better suited for handling varied object scales and complex road scenes \cite{law2018cornernet}.

\subsubsection{CenterNet}
CenterNet, introduced in the 2019 work \emph{Objects as Points}, pushed anchor-free detection further by representing each object as a center point. Its known innovation was simplifying detection to center heatmap prediction plus regression of object properties such as size, offset, and even 3D attributes in certain settings. This reduced the need for anchor adaptations and heavy post-processing. 

Architecturally, CenterNet predicts center heatmaps and regresses width-height and local offsets from shared features. Its loss blends heatmap-based classification with regression terms for geometric properties. While the detector can be elegant and fast, it can still face difficulty in crowded scenes where centers are close together, and performance depends heavily on feature quality. For autonomous driving, its clean center-based formulation is appealing because road-scene objects often benefit from multi-task geometric prediction \cite{zhou2019objects}.

\subsubsection{FCOS}
FCOS was introduced in 2019 as a fully convolutional one-stage detector that predicts boxes per pixel. Its main contribution was eliminating anchor boxes and proposal generation entirely, replacing them with per-location predictions and centerness estimation. This simplified the detection pipeline and reduced hyperparameter sensitivity tied to anchor design.

FCOS uses a backbone and FPN-style multi-level features, followed by shared heads that output class, box regression, and centerness predictions. Its losses are designed around dense classification and localization without anchors. Although FCOS improves simplicity and often performs strongly, it still struggles with ambiguous assignments in complex scenes. In autonomous driving, it is especially relevant because anchor-free design can simplify adaptation to varied road-object scales and aspect ratios \cite{tian2019fcos}

\subsection{Modern Dense and Efficiency-Oriented Detectors}
Recent one-stage detectors have increasingly focused on improving localization quality, confidence calibration, multi-scale feature fusion, and deployment efficiency. These developments are especially important for autonomous driving, where perception systems must satisfy both accuracy and real-time constraints \cite{lin2017focalloss,tan2020efficientdet,li2020gfl,zhang2021vfnet,lyu2022rtmdet}.

\subsubsection{RetinaNet}
RetinaNet was introduced in 2017 and became one of the most crucial dense detectors because it addresses the foreground-background imbalance problem. Its most notable contribution was the focal loss, which down-weights easy negatives so the model focuses on hard examples during training.

RetinaNet uses a backbone with a feature pyramid network and separate subnetworks for classification and box regression. Its classification branch is trained with Focal Loss, while localization is optimized separately for bounding boxes. Despite its strong accuracy, RetinaNet can be heavier than some real-time YOLO style models, so its deployment in AD may depend on the hardware constraints \cite{lin2017focalloss}.

\subsubsection{EfficientDet}
EfficientDet was introduced in 2020 as a scalable and efficient detector family. Its two biggest contributions were the BiFPN for weighted bidirectional multi-scale feature fusion and compound scaling, which jointly scales backbone, feature network, head, and image resolution. This in total made the detector family adaptable.

EfficientDet builds EfficientNet backbones, BiFPN feature fusion, and lightweight prediction heads. The design uses efficient modules such as depthwise separable convolutions while still being able to optimize standard dense detection objectives. Its main limitation is that the family is difficult to customize, but its usefulness comes when it performs on embedded platforms and onboard accelerators \cite{tan2020efficientdet}. 

\subsubsection{GFL}
Generalized Focal Loss was introduced in 2020 to address inconsistencies between classification confidence and localization quality in dense detection. Its innovation was learning a joint representation of class confidence and localization quality while also modeling box locations as distributions rather than fixed point estimates. This made dense detectors better at ranking detections and expressing localization uncertainty. 

Architecturally, GFL is typically built on a standard dense detector framework but changes the prediction representation and loss design. The loss unifies classification-quality estimation and uses a distributional view of bounding-box regression. Its limitation is that the method is more careful than architectural, so its benefits depend on a strong base detector and careful implementation. For autonomous driving, GFL is useful because better ranking and uncertainty-aware localization matter in safety-critical scenes \cite{li2020gfl}.

\subsubsection{VFNet}
VFNet was introduced in 2021 as an IoU-aware dense detector. Its central contribution was the IoU-Aware Classification Score and Varifocal Loss, which aims to align classification confidence more closely with localization quality so that final ranking is more reliable. This directly targets a persistent weakness in one-stage detection: high class scores do not always correspond to well-localized boxes. 

Architecturally, VFNet adds mechanisms for quality-aware scoring and box refinement on top of dense detection features. Its loss and scoring design are the main novelty rather than a radical new backbone. The detector improves ranking quality and accuracy, but it also introduces additional design complexity. In autonomous driving, VFNet is relevant because it is reliable and better localization confidence can improve decisions when multiple road objects compete for attention \cite{zhang2021vfnet}.

\subsubsection{RTMDet}
RTMDet was introduced in 2022 as a real-time detector designed through systematic empirical study. Its main contribution was a balanced backbone-neck design built around efficient large-kernel depthwise convolutions, plus improved training choices such as soft-label matching costs.

Architecturally, RTMDet emphasizes compatibility between model stages and strong performance across several scales. Its training and assignment refinements help improve accuracy while keeping high throughput. A limitation is that it is newer and less historically foundational than earlier detectors, so in a survey it is best framed as a modern endpoint in the evolution of real-time one-stage design. In autonomous driving, RTMDet fits well because it directly targets the speed-accuracy-efficiency balance that onboard perception systems need \cite{lyu2022rtmdet}

\section{Benchmark Datasets}

Benchmark datasets are important because they define the conditions under which one-stage detectors are evaluated. In autonomous driving, datasets differ widely in sensor setup, object classes, scene diversity, annotation style, and evaluation protocol. Table~\ref{tab:datasets} summarizes commonly used datasets and shows that no single benchmark fully captures all deployment conditions. KITTI is historically important and widely used, but it is smaller than newer datasets. Waymo, nuScenes, BDD100K, and Argoverse provide broader coverage of complex road scenes, while COCO is often used for pretraining even though it is not specific to autonomous driving. Because of these differences, performance values should be interpreted with the dataset context in mind rather than treated as directly interchangeable.

\begin{table*}[htbp]
\caption{Benchmark datasets used in autonomous driving object detection}
\label{tab:datasets}
\centering
\scriptsize
\renewcommand{\arraystretch}{1.2}
\setlength{\tabcolsep}{4pt}
\begin{tabularx}{\textwidth}{L{0.13\textwidth} c c c Y L{0.16\textwidth}}
\toprule
\textbf{Dataset} & \textbf{Year} & \textbf{\# Images} & \textbf{\# Classes} & \textbf{Key AV Focus / Characteristics} & \textbf{Commonly Used Metric} \\
\midrule

KITTI
& 2012
& 7,481 train / 7,518 test
& 8
& Urban driving benchmark with cars, pedestrians, cyclists, vans, trucks, trams, and other road objects. Often used for 2D/3D detection, depth, tracking, and distance estimation.
& AP, mAP, 3D AP \\[0.8em]

Waymo Open Dataset
& 2019
& 1M+ camera images
& 4 main detection classes
& Large-scale real-world driving dataset with camera and LiDAR data across diverse road scenes. Useful for evaluating detection in complex traffic environments.
& mAP, mAPH \\[0.8em]

nuScenes
& 2019
& 1.4M camera images
& 10 detection classes
& Multi-sensor dataset with cameras, LiDAR, radar, GPS, and IMU. Strong for autonomous driving because it evaluates detection across full 360-degree driving scenes.
& mAP, NDS \\[0.8em]

BDD100K
& 2020
& 100,000 images
& 10 detection classes
& Large driving dataset with diverse weather, lighting, road types, and time-of-day conditions. Useful for studying real-world robustness.
& mAP \\[0.8em]

Cityscapes
& 2016
& 5,000 finely annotated images
& 8 instance-level classes
& Urban street-scene dataset focused on dense pixel-level understanding. Often used for segmentation, but still useful for evaluating road-scene perception quality.
& mIoU, AP \\[0.8em]

Argoverse
& 2019
& 290,000+ images
& 15
& Autonomous driving dataset focused on 3D tracking, maps, and urban driving scenes. Useful for detection and motion understanding in city environments.
& mAP, tracking metrics \\[0.8em]

COCO
& 2017
& 123,000 images
& 80
& General object detection benchmark often used for pretraining one-stage detectors before fine-tuning on autonomous-driving datasets.
& COCO mAP \\

\bottomrule
\end{tabularx}
\end{table*}

\begin{figure*}[!t]
\centering
\includegraphics[width=0.88\textwidth]{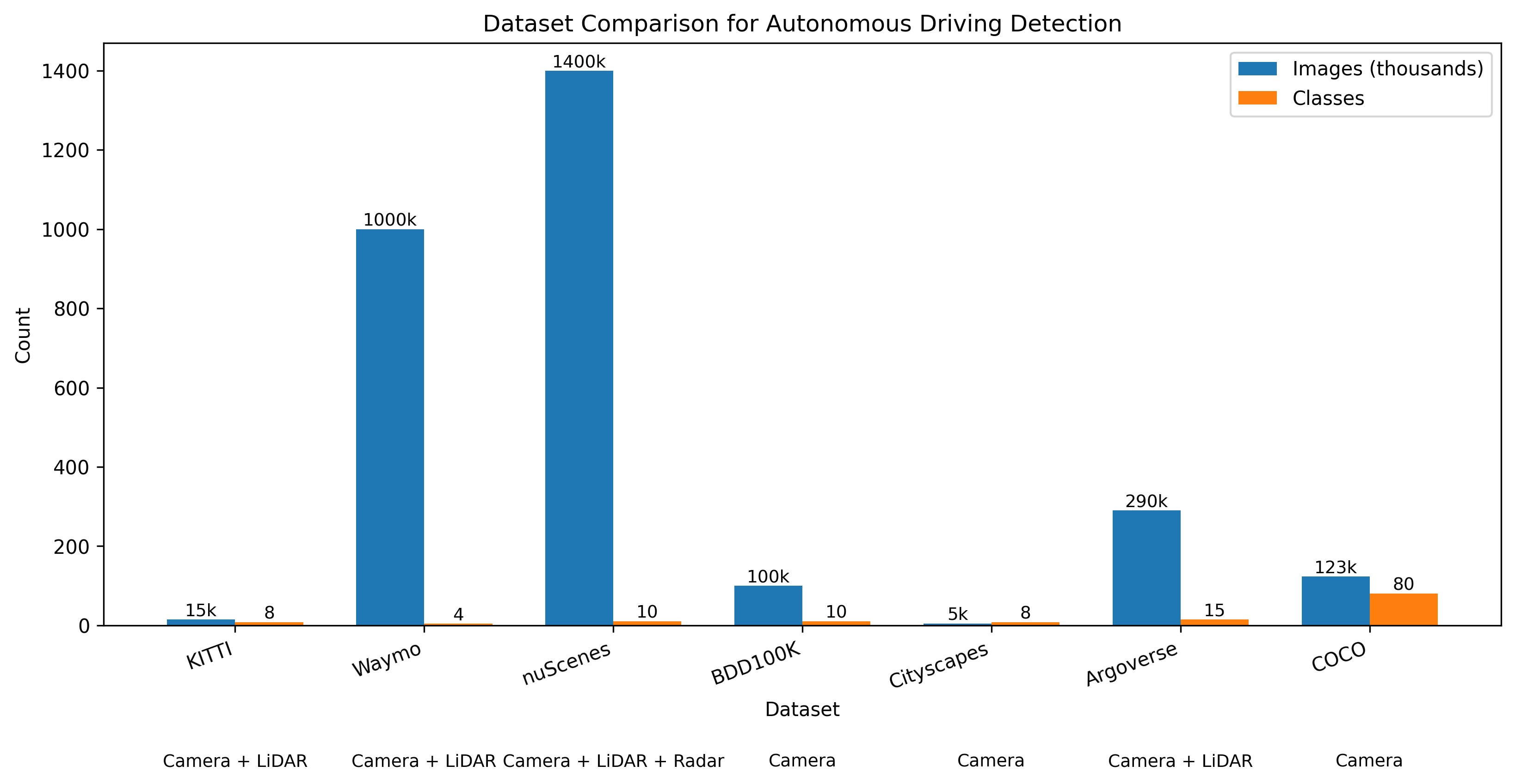}
\caption{Comparison of major benchmark datasets used for autonomous driving detection.}
\label{fig:dataset_comparison}
\end{figure*}

\section{Evaluation Metrics}

Evaluation metrics provide different views of detector performance. Accuracy-focused metrics such as AP, mAP, IoU, precision, and recall describe how well the model detects and localizes objects. Deployment-focused metrics such as FPS, latency, parameter count, and FLOPs describe whether the model can run within real-time and hardware constraints. For autonomous driving, both categories matter: a detector with high mAP may still be impractical if its latency is too high, while a very fast detector may be unsafe if it misses small or distant road users. Table~\ref{tab:metrics} summarizes the metrics used throughout this survey and explains how each metric should be interpreted.

\begin{table}[htbp]
\caption{Evaluation metrics used in this survey}
\label{tab:metrics}
\centering
\scriptsize
\renewcommand{\arraystretch}{1.18}
\setlength{\tabcolsep}{3pt}
\begin{tabularx}{\columnwidth}{L{0.18\columnwidth} L{0.23\columnwidth} Y L{0.16\columnwidth}}
\toprule
\textbf{Metric} & \textbf{Full Name} & \textbf{What It Measures} & \textbf{Higher is Better?} \\
\midrule

mAP
& mean Average Precision
& Shows how well the detector finds objects and labels them correctly across all classes.
& Yes \\[0.8em]

AP
& Average Precision
& Measures how well the detector finds one specific object class, such as cars or pedestrians.
& Yes \\[0.8em]

IoU
& Intersection over Union
& Checks how much the predicted box overlaps with the real object box.
& Yes \\[0.8em]

FPS
& Frames Per Second
& Shows how many images the model can process each second.
& Yes \\[0.8em]

Latency
& Inference Delay
& Measures how long the model takes to make one prediction.
& No \\[0.8em]

Parameters
& Model Parameters
& Shows how large the model is based on how many learned values it has.
& No \\[0.8em]

FLOPs
& Floating-Point Operations
& Estimates how much computation the model needs to process an image.
& No \\[0.8em]

Recall
& Recall
& Shows how many real objects the detector successfully finds.
& Yes \\[0.8em]

Precision
& Precision
& Shows how many predicted detections are actually correct.
& Yes \\[0.8em]

RMSE
& Root Mean Square Error
& Shows how far the model's predictions are from the correct values on average.
& No \\

\bottomrule
\end{tabularx}
\end{table}

\section{Performance Comparison}

Table~\ref{tab:performance} compares reported performance values for the selected detectors. These values should be read as a survey-style comparison rather than a controlled benchmark because the original papers use different datasets, input sizes, hardware platforms, and reporting conventions. For this reason, missing values are marked as N/R, meaning that the value was not reported in the referenced paper or was not reported in a directly comparable way.

\begin{table*}[htbp]
\caption{Performance comparison of one-stage detectors. Values are reported from the original papers when available. Because the papers use different datasets, metrics, input sizes, and hardware, the results should be interpreted as reported benchmark values rather than a perfectly controlled comparison.}
\label{tab:performance}
\centering
\scriptsize
\renewcommand{\arraystretch}{1.15}
\setlength{\tabcolsep}{3pt}

\begin{tabularx}{\textwidth}{L{0.11\textwidth} c L{0.15\textwidth} L{0.12\textwidth} c c c c}
\toprule
\textbf{Detector} & \textbf{Year} & \textbf{Backbone} & \textbf{Dataset} & \textbf{mAP/AP (\%)} & \textbf{FPS} & \textbf{Params (M)} & \textbf{Input Size} \\
\midrule

YOLOv1 \cite{redmon2016yolo}
& 2016
& Custom CNN
& VOC 2007
& 63.4
& 45
& N/R
& 448 $\times$ 448 \\[0.7em]

YOLOv2 \cite{redmon2017yolo9000}
& 2017
& Darknet-19
& VOC 2007
& 76.8
& 67
& N/R
& 416 $\times$ 416 \\[0.7em]

YOLOv3 \cite{redmon2018yolov3}
& 2018
& Darknet-53
& COCO
& 33.0
& 20
& N/R
& 608 $\times$ 608 \\[0.7em]

YOLOv4 \cite{bochkovskiy2020yolov4}
& 2020
& CSPDarknet53
& COCO
& 43.5
& 62
& 64
& 608 $\times$ 608 \\[0.7em]

YOLOX \cite{ge2021yolox}
& 2021
& CSPDarknet
& COCO
& 50.0
& 68.9
& 54
& 640 $\times$ 640 \\[0.7em]

YOLOv7 \cite{wang2023yolov7}
& 2023
& E-ELAN
& COCO
& 51.4
& 161
& 36.9
& 640 $\times$ 640 \\[0.7em]

YOLOv10 \cite{wang2024yolov10}
& 2024
& CSPNet-based
& COCO
& 46.3
& N/R
& 7.2
& 640 $\times$ 640 \\[0.7em]

SSD \cite{liu2016ssd}
& 2016
& VGG-16
& VOC 2007
& 72.1
& 58
& N/R
& 300 $\times$ 300 \\[0.7em]

DSSD \cite{fu2017dssd}
& 2017
& ResNet-101
& VOC 2007
& 78.6
& N/R
& N/R
& 321 $\times$ 321 \\[0.7em]

RefineDet \cite{zhang2018refinedet}
& 2018
& VGG-16
& VOC 2007
& 80.1
& N/R
& N/R
& 512 $\times$ 512 \\[0.7em]

CornerNet \cite{law2018cornernet}
& 2018
& Hourglass-104
& COCO
& 42.1
& 4.1
& N/R
& 511 $\times$ 511 \\[0.7em]

CenterNet \cite{zhou2019objects}
& 2019
& DLA-34
& COCO
& 37.4
& 52
& N/R
& 512 $\times$ 512 \\[0.7em]

FCOS \cite{tian2019fcos}
& 2019
& ResNeXt-101-FPN
& COCO
& 44.7
& N/R
& N/R
& 800 $\times$ 1024 \\[0.7em]

RetinaNet \cite{lin2017focalloss}
& 2017
& ResNet-101-FPN
& COCO
& 39.1
& 5
& N/R
& 800 $\times$ 1333 \\[0.7em]

EfficientDet-D0 \cite{tan2020efficientdet}
& 2020
& EfficientNet-B0 + BiFPN
& COCO
& 33.8
& 134
& 3.9
& 512 $\times$ 512 \\[0.7em]

GFL \cite{li2020gfl}
& 2020
& ResNet-101
& COCO
& 45.0
& N/R
& N/R
& 800 $\times$ 1333 \\[0.7em]

VFNet \cite{zhang2021vfnet}
& 2021
& Res2Net-101-DCN
& COCO
& 55.1
& 4.2
& N/R
& 1200 $\times$ 1200 \\[0.7em]

RTMDet \cite{lyu2022rtmdet}
& 2022
& CSPNeXt
& COCO
& 52.8
& 300+
& N/R
& 640 $\times$ 640 \\

\bottomrule
\end{tabularx}
\end{table*}

\begin{figure*}[!t]
\centering
\includegraphics[width=0.88\textwidth]{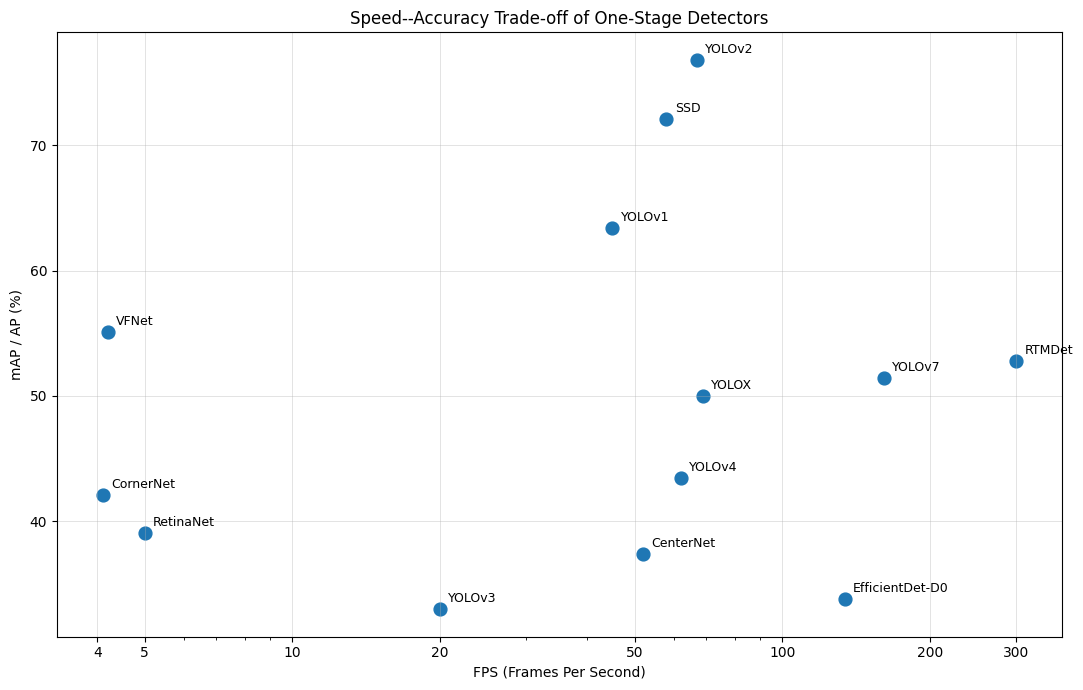}
\caption{Speed-accuracy trade-off plot of one-stage detectors.}
\label{fig:speed_accuracy}
\end{figure*}

\section{Advantages, Disadvantages and Limitations}

This section compares the major strengths, weaknesses, and deployment limitations of the selected one-stage detectors. While many one-stage models are designed for real-time inference, their practical usefulness in autonomous driving depends on more than speed alone. Factors such as small-object detection, robustness under occlusion, computational cost, and suitability for edge hardware all affect whether a detector can be reliably deployed in an AV perception pipeline. Table~\ref{tab:adl} summarizes these trade-offs by showing how each detector family performs from both an algorithmic and deployment-focused perspective.

\begin{table*}[!t]
\caption{Advantages, disadvantages, and deployment limitations of one-stage detectors}
\label{tab:adl}
\centering
\scriptsize
\renewcommand{\arraystretch}{1.18}
\setlength{\tabcolsep}{4pt}

\begin{tabularx}{\textwidth}{L{0.14\textwidth} Y Y Y}
\toprule
\textbf{Detector} & \textbf{Key Advantages} & \textbf{Key Disadvantages} & \textbf{AV Deployment Limitation} \\
\midrule

YOLO (v4 / v8 / v10 family)
& Very high inference speed, making it real time capable. single-stage architecture. strong speed and accuracy balance. scalable (small can turn into large variants). efficient for edge deployment.
& Slightly lower localization accuracy than two-stage models. struggles with very small or occluded objects. more performance sensitive to dataset quality than other models.
& May miss small/distant objects (e.g., pedestrians, bikes) in dense traffic. requires careful tuning for safety critical scenarios, such as unique problem cases. \\[0.8em]

SSD (Single Shot Detector)
& Fast inference (often the highest FPS of all one-stage models). Lightweight. Simple architecture. suitable for embedded/low-power devices. More efficient for computers.
& Lower accuracy compared to YOLO and EfficientDet. weak performance on small objects. limited feature representation depth.
& High miss rate for small/critical objects which reduces reliability in AV perception pipelines. \\[0.8em]

RetinaNet
& Good balance between accuracy and handling class imbalance (Focal Loss). strong detection of hard/rare objects.
& Slower than single-stage models like YOLO and SSD. computationally heavier. still struggles with latency constraints.
& Difficult to meet real-time AV requirements without optimization. limited edge deployment feasibility. \\[0.8em]

RefineDet
& Improves SSD with two-step refinement. better localization accuracy than SSD. maintains relatively high speed. Multi-stage refinement improves detection quality.
& More complex than SSD. slower than pure one-stage models. still weaker than YOLO in speed; increased training complexity.
& May not meet strict real-time latency for high-resolution AV inputs. harder to deploy on low-power edge hardware. \\[0.8em]

CornerNet
& Anchor-free detection (predicts object corners). Reduces dependence on anchor tuning. Good localization for irregular shapes.
& Computationally expensive. slow inference. Grouping corners is complex. sensitive to detection errors at corners.
& Not suitable for real-time AV systems. latency too high for onboard perception pipelines. \\[0.8em]

CenterNet
& Detect objects via center points. simpler than corner-based methods. A good balance between accuracy and speed. fewer hyperparameters (no anchors).
& Still slower than YOLO. struggles with dense object clustering. center-point ambiguity in crowded scenes.
& May miss closely packed objects (e.g., urban traffic). borderline real-time without optimization. \\[0.8em]

FCOS
& Anchor-free. simpler training pipeline. strong performance on small objects. competitive accuracy with good efficiency. flexible across backbones.
& Slightly slower than YOLO. sensitive to feature pyramid quality. Still requires tuning for optimal performance.
& May not consistently meet strict latency constraints on edge AV hardware. Requires optimization for deployment. \\[0.8em]

EfficientDet
& Efficient scaling with BiFPN and compound scaling. strong balance between accuracy and parameter efficiency. good for resource-aware detection.
& Can be slower at larger model scales. training and tuning can be complex. performance depends heavily on selected model scale.
& Useful for edge-aware AV systems, but larger variants may exceed latency or compute limits. \\

\bottomrule
\end{tabularx}
\end{table*}

\FloatBarrier

\subsection{Cross-Cutting Patterns and Observations}

Across all detectors, a clear pattern emerges: no single model simultaneously optimizes accuracy, speed, and deployment feasibility. Instead, each detector occupies a different point along the trade-off spectrum. Single-stage detectors such as YOLO and SSD dominate real-time performance, making them the most viable for AV deployment, particularly in latency-sensitive tasks. However, this speed advantage often comes at the cost of reduced accuracy in challenging scenarios, especially for small, distant, or occluded objects.

Another important trend is that deployment constraints extend beyond FPS. While many studies emphasize inference speed, practical AV deployment also depends on hardware limitations such as memory, power, embedded computing resources, robustness to environmental variability, and system integration complexity. For example, SSD may achieve the highest FPS, but its weaker detection capability for small objects makes it less suitable for safety-critical perception. Similarly, EfficientDet offers strong accuracy but becomes impractical at higher scales due to latency.

Overall, the literature highlights a gap between benchmark performance and real-world deployment readiness. Models that perform well in controlled experiments may fail under real-world conditions such as poor lighting, weather variability, or dense traffic. This suggests that future AV perception systems must move toward hardware-aware, robustness-driven, and system-level optimized models, rather than relying solely on improvements in isolated detection accuracy or speed.

When incorporating newer one-stage detectors, a broader pattern becomes clear: anchor-free methods such as CornerNet, CenterNet, and FCOS aim to simplify detection pipelines and remove the need for anchor box tuning, which improves training stability and generalization. However, this simplification does not automatically translate to real-time deployment benefits. In fact, some anchor-free approaches, such as CornerNet, introduce computational overhead due to complex keypoint grouping, making them unsuitable for AV latency requirements despite their theoretical advantages.

Compared to these newer detectors, YOLO remains the most deployment-ready architecture due to its consistent emphasis on speed optimization and architectural efficiency. Models like FCOS and CenterNet offer promising alternatives with competitive accuracy and simpler design, but they still require hardware-aware optimization to match YOLO's real-time performance in AV systems.

Overall, the extended comparison reinforces a key insight: real-time AV deployment favors detectors that are explicitly designed for low latency, not just high accuracy or architectural elegance. Future work must bridge the gap between algorithmic innovation, such as anchor-free detection, and practical deployment constraints, ensuring that improvements translate into measurable gains in real-world AV perception systems \cite{yilmaz2025realtime,balasubramaniam2022status,du2023object}.

\FloatBarrier

\begin{figure*}[!t]
\centering
\includegraphics[width=0.80\textwidth]{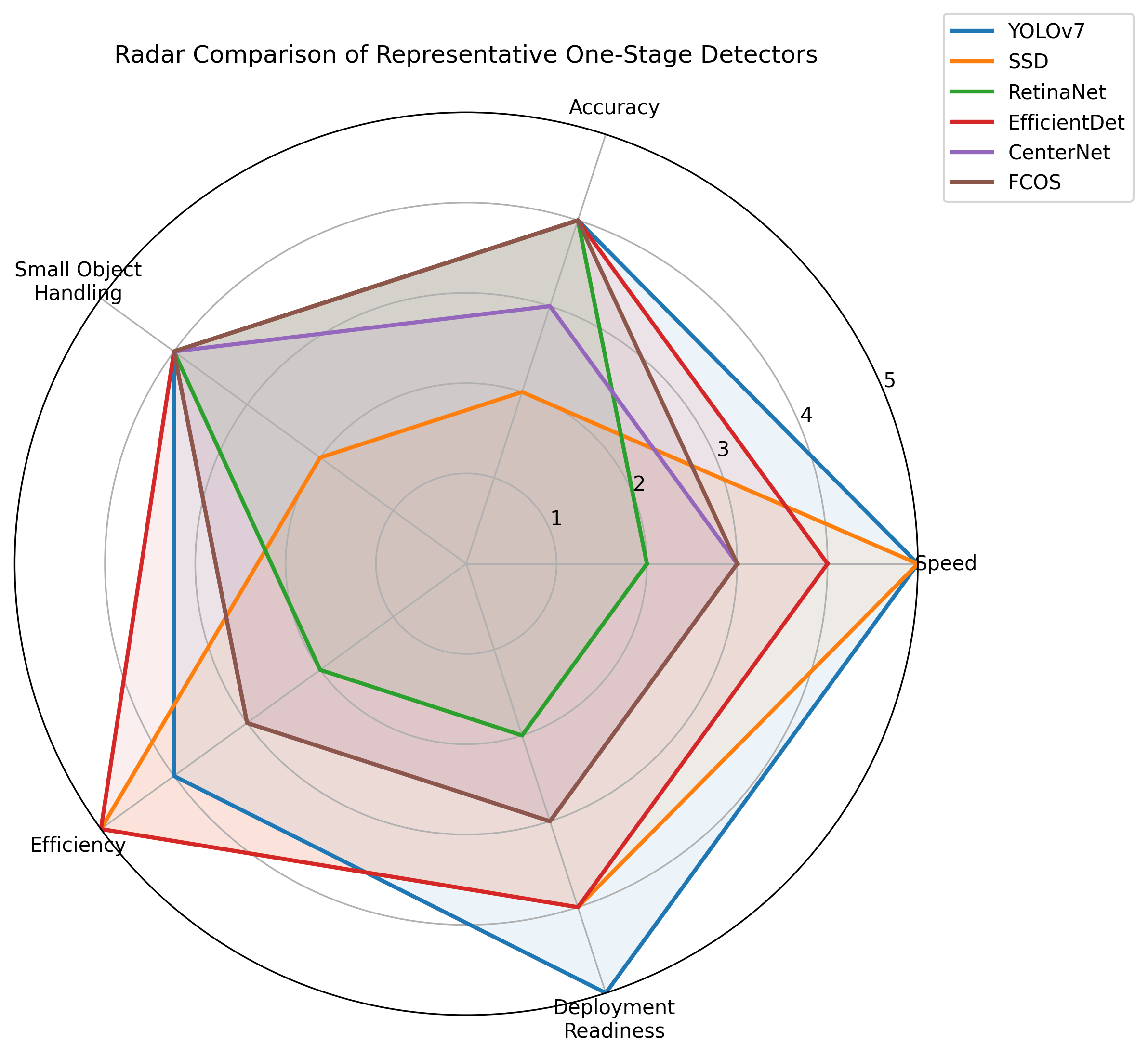}
\caption{Radar chart comparing representative one-stage detectors across multiple dimensions.}
\label{fig:radar}
\end{figure*}

The radar chart in Fig.~\ref{fig:radar} uses a qualitative five-point scale to compare representative detectors across five deployment-oriented dimensions. Accuracy reflects the reported detection quality from the original studies, using mAP/AP as the main guide. Speed reflects reported FPS or latency where available. Efficiency considers parameter count, computational cost, and whether the model was designed for resource-aware deployment. Deployment readiness reflects how practical the detector appears for real-time autonomous-driving use, including ecosystem support and the need for post-processing. Small-object handling reflects the detector's use of multi-scale features, feature pyramids, or design choices that improve detection of distant pedestrians, cyclists, signs, and other small road objects. These scores are therefore not a new benchmark result. Instead, they are a survey-based summary derived from the reported properties and limitations of each model.

\section{Open Challenges in One-Stage Detection for AVs}

Although one-stage detectors are strong candidates for autonomous-driving perception, several challenges remain unresolved. Table~\ref{tab:challenges} summarizes the main limitations that appear across the surveyed literature. The most important pattern is that benchmark progress does not automatically guarantee safe deployment: a detector may perform well on common datasets while still failing on small, distant, occluded, or weather-degraded objects. Autonomous driving also introduces strict hardware and latency constraints, so future detectors must be evaluated not only by mAP, but also by real-time performance, robustness, and system-level readiness.

\begin{table}[!t]
\caption{Open challenges in one-stage detection for autonomous driving}
\label{tab:challenges}
\centering
\scriptsize
\renewcommand{\arraystretch}{1.12}
\setlength{\tabcolsep}{4pt}

\begin{tabularx}{\columnwidth}{L{0.22\columnwidth} Y Y Y}
\toprule
\textbf{Challenge} & \textbf{Description} & \textbf{Why It Matters for AVs} & \textbf{Partially Addressed By} \\
\midrule
Speed vs. Accuracy Trade-off
& Faster models sacrifice precision, especially for small or complex objects
& Missed or incorrect detections can lead to unsafe driving decisions
& YOLO (balanced), EfficientDet (scaling trade-off) \\

Small Object Detection \& Occlusion
& Difficulty detecting small, distant, or partially hidden objects
& Critical objects like pedestrians or bikes may be missed
& YOLO (multi-scale), FCOS (feature pyramids), CenterNet \\

Robustness to Environmental Conditions
& Performance drops under poor lighting, weather, and sensor noise
& AVs must operate reliably in all conditions for safety
& YOLOv5 (robustness improvements), data augmentation methods \\

Resource-Constrained Deployment
& Limited compute, memory, and power on edge devices
& Prevents real-time inference or requires heavy optimization
& SSD (lightweight), YOLOv5/YOLOv8 (efficient variants) \\

Lack of Deployment-Centric Metrics
& Over-reliance on mAP and FPS ignores real-world constraints
& Misleading evaluation of model readiness for AV systems
& EfficientDet (efficiency-aware design), recent benchmarking studies \\

Anchor-Free Localization Limitations
& Anchor-free models struggle in dense or overlapping scenes
& Incorrect localization affects tracking and decision-making
& FCOS, CenterNet (partial solutions) \\

\bottomrule
\end{tabularx}
\end{table}

\FloatBarrier

\subsection{Performance on Autonomous-Driving-Specific Data}

A major issue in comparing one-stage detectors is that many well-known results are reported on general-purpose datasets such as COCO or Pascal VOC, while autonomous driving requires reliable performance on road-scene datasets such as KITTI, Waymo, nuScenes, BDD100K, Cityscapes, and Argoverse. Performance on these datasets is more relevant to AV deployment because the objects of interest are safety-critical and often appear under difficult conditions, including long distance, partial occlusion, nighttime lighting, rain, fog, dense traffic, and unusual viewpoints. Studies focused on autonomous driving show that models such as YOLO, SSD, RetinaNet, EfficientDet, FCOS, and related one-stage methods must be judged by how well they handle vehicles, pedestrians, cyclists, traffic signs, and other road users rather than by general-object performance alone \cite{arnold2021onperformance,balasubramaniam2022status,du2023object,liang2024vehicle}.

In practice, this means that the strongest detector for a general benchmark is not always the most suitable detector for an AV pipeline. YOLO-style detectors are often attractive because they offer high FPS and practical deployment options, but they may still miss small or distant objects without careful training and multi-scale design. SSD is lightweight and fast, but its small-object weakness is a serious concern for pedestrians, cyclists, and road signs. RetinaNet, FCOS, EfficientDet, and RTMDet improve different parts of the accuracy-efficiency trade-off, but they must still be validated on driving-specific datasets and hardware before being considered deployment-ready. For this reason, future comparisons should report AV-specific results alongside general benchmark results, including per-class performance for cars, pedestrians, cyclists, and traffic signs, as well as results under adverse weather and low-light conditions.

\section{Future Directions}

\subsection{Improving Small Object Detection in Road Scenes}

One important future direction is improving how one-stage detectors handle small and distant objects. In autonomous driving, objects such as pedestrians, cyclists, traffic signs, and motorcycles may appear very small in the camera image, especially when they are far away. If the detector misses these objects, the vehicle may not have enough time to react safely. Future research should focus on stronger multi-scale feature extraction, better feature fusion, and training methods that give more attention to small objects. This would help AV perception systems detect important road users earlier and more reliably \cite{mahadshetti2024signyolo,chan2023multispectral}.

\subsection{Building More Robust Models for Bad Weather}

Another future direction is making object detectors more reliable in bad weather and poor lighting. Many models perform well on clear images but lose accuracy in rain, fog, haze, glare, or nighttime conditions. Since autonomous vehicles must operate in real-world environments, detectors should be tested and trained on more difficult weather conditions. Future work could combine image enhancement with object detection, so the model can improve the image while also focusing on the detection task. This would make AV systems safer because the perception module would be less dependent on perfect weather or lighting \cite{lee2022badweather}.

\subsection{Designing Faster Models for Edge Hardware}

A major challenge for autonomous driving is running detection models on limited onboard hardware. A model may have high accuracy in testing, but it may not be useful if it is too slow or too large for real-time deployment. Future research should focus on lightweight model design, pruning, quantization, and hardware optimization. These methods can reduce model size and inference time while keeping enough accuracy for safety. This direction is important because AVs need fast decisions without relying on unlimited computing power \cite{tan2020efficientdet,lyu2022rtmdet}.

\subsection{Creating Better Real-World Evaluation Metrics}

Future work should also improve how one-stage detectors are evaluated. Metrics such as mAP and FPS are useful, but they do not fully show whether a detector is ready for autonomous driving. A model could have strong benchmark results but still fail in dense traffic, bad weather, or rare safety-critical situations. Future evaluation should include latency, missed detections of critical objects, performance under weather changes, and hardware usage. This would give a more realistic picture of whether a detector is truly reliable for AV deployment \cite{arnold2021onperformance,balasubramaniam2022status}.

\subsection{Connecting Detection with the Full AV Pipeline}

Another future direction is connecting object detection more closely with the rest of the autonomous driving system. Detection is only one part of the AV pipeline, and its results affect tracking, localization, planning, and control. Future models should not only predict bounding boxes and labels, but also provide information that helps the vehicle make safer driving decisions. For example, detectors could include uncertainty estimates, object distance, or motion information. This would help the vehicle understand not only what objects are present, but also how risky each situation may be \cite{liu2023lidarcamera,liu2024largeai}.

\subsection{Improving Anchor-Free Detection for Real-Time Use}

Anchor-free detectors such as CornerNet, CenterNet, and FCOS simplify the detection process by removing anchor box tuning. However, some of these methods still have speed or localization problems, especially in crowded road scenes. Future research should focus on making anchor-free detectors faster and more stable for real-time AV use. This could include simpler prediction heads, better center-point or keypoint matching, and improved handling of overlapping objects. If these models become faster and more reliable, they could become stronger alternatives to YOLO-style detectors in autonomous driving \cite{law2018cornernet,zhou2019objects,tian2019fcos}.
\section{Conclusion}

The continuous evolution of one-stage detectors has shown a major shift from simple, speed-focused architectures to more advanced models that try to balance real-time efficiency with accurate detection. By using improvements such as better feature fusion, stronger loss functions, and more efficient model designs, these detectors address many of the trade-offs that matter in autonomous driving. The research suggests that reliable object detection depends not only on the detector family being used, but also on how well the model handles speed, accuracy, localization, and real-world driving conditions. Moving forward, the field still needs to improve small-object detection, occlusion handling, bad-weather performance, and deployment on limited onboard hardware. Overall, this review shows that one-stage detectors remain an important part of autonomous vehicle perception, but further work is needed before they can be fully dependable in real-world driving environments \cite{balasubramaniam2022status,lee2022badweather,yilmaz2025realtime}.


\begin{thebibliography}{00}

\bibitem{yilmaz2025realtime}
E. N. Yilmaz and T. S. Navruz, ``Real-Time Object Detection: A Comparative Analysis of YOLO, SSD, and EfficientDet Algorithms,'' in \emph{Proc. 2025 7th International Congress on Human-Computer Interaction, Optimization and Robotic Applications}, Ankara, Turkey, 2025, pp. 1--9.

\bibitem{yinkfu2025motorbike}
N. Yinkfu, S. Nwovu, J. Kayizzi, and A. Uwamahoro, ``Comparative Analysis of YOLOv5, Faster R-CNN, SSD, and RetinaNet for Motorbike Detection in Kigali Autonomous Driving Context,'' \emph{arXiv preprint}, 2025.

\bibitem{pothole2025yolov5ssd}
``Robust Pothole Detection Using YOLOv5 and SSD: Accuracy, Speed, and Deployment Trade-offs,'' IEEE Xplore, 2025.

\bibitem{sarda2021yolo}
A. Sarda, S. Dixit, and A. Bhan, ``Object Detection for Autonomous Driving using YOLO You Only Look Once Algorithm,'' in \emph{Proc. 2021 Third International Conference on Intelligent Communication Technologies and Virtual Mobile Networks}, Tirunelveli, India, 2021, pp. 1370--1374.

\bibitem{liu2023lidarcamera}
H. Liu, C. Wu, and H. Wang, ``Real Time Object Detection Using LiDAR and Camera Fusion for Autonomous Driving,'' \emph{Scientific Reports}, vol. 13, no. 1, 2023.

\bibitem{arnold2021onperformance}
E. Arnold, O. Y. Al-Jarrah, M. Dianati, S. Fallah, D. Oxtoby, and A. Mouzakitis, ``On the Performance of One-Stage and Two-Stage Object Detectors in Autonomous Vehicles Using Camera Data,'' \emph{Remote Sensing}, vol. 13, no. 1, Art. no. 89, 2021.

\bibitem{tan2020efficientdet}
M. Tan, R. Pang, and Q. V. Le, ``EfficientDet: Scalable and Efficient Object Detection,'' in \emph{Proc. IEEE/CVF Conference on Computer Vision and Pattern Recognition}, 2020, pp. 10781--10790.

\bibitem{feng2021tood}
C. Feng, Y. Zhong, Y. Gao, M. R. Scott, and W. Huang, ``TOOD: Task-Aligned One-Stage Object Detection,'' \emph{arXiv preprint arXiv:2108.07755}, 2021.

\bibitem{lyu2022rtmdet}
C. Lyu, W. Zhang, H. Huang, Y. Zhou, Y. Wang, Y. Liu, S. Zhang, and K. Chen, ``RTMDet: An Empirical Study of Designing Real-Time Object Detectors,'' \emph{arXiv preprint arXiv:2212.07784}, 2022.

\bibitem{xu2022ppyoloe}
S. Xu, X. Wang, W. Lv, Q. Chang, C. Cui, K. Deng, G. Wang, Q. Dang, S. Wei, Y. Du, and B. Lai, ``PP-YOLOE: An Evolved Version of YOLO,'' \emph{arXiv preprint arXiv:2203.16250}, 2022.

\bibitem{zeng2020refpn}
J. Zeng, J. Xiong, X. Fu, and L. Leng, ``ReFPN-FCOS: One-Stage Object Detection for Feature Learning and Accurate Localization,'' \emph{IEEE Access}, vol. 8, pp. 225052--225063, 2020.

\bibitem{ieee9293286}
J. Fan, T. Huo, and X. Li, ``A Review of One-Stage Detection Algorithms in Autonomous Driving,'' in \emph{Proc. 2020 4th CAA International Conference on Vehicular Control and Intelligence}, Hangzhou, China, 2020, pp. 210--214, doi: 10.1109/CVCI51460.2020.9338663.

\bibitem{zhao2018dynamic}
K. Zhao, X. Zhu, H. Jiang, C. Zhang, Z. Wang, and B. Fu, ``Dynamic Loss for One-Stage Object Detectors in Computer Vision,'' \emph{Electronics Letters}, vol. 54, no. 25, pp. 1433--1434, 2018.

\bibitem{du2023object}
L. Du, ``Object Detectors in Autonomous Vehicles: Analysis of Deep Learning Techniques,'' \emph{International Journal of Advanced Computer Science and Applications}, vol. 14, no. 10, pp. 217--224, 2023.

\bibitem{sirisha2023yolo}
U. Sirisha, S. P. Praveen, P. N. Srinivasu, and others, ``Statistical Analysis of Design Aspects of Various YOLO-Based Deep Learning Models for Object Detection,'' \emph{International Journal of Computational Intelligence Systems}, vol. 16, Art. no. 126, 2023.

\bibitem{mittal2020uav}
P. Mittal, R. Singh, and A. Sharma, ``Deep Learning-Based Object Detection in Low-Altitude UAV Datasets: A Survey,'' \emph{Image and Vision Computing}, vol. 104, Art. no. 104046, 2020.

\bibitem{liang2024vehicle}
L. Liang, H. Ma, L. Zhao, X. Xie, C. Hua, M. Zhang, and Y. Zhang, ``Vehicle Detection Algorithms for Autonomous Driving: A Review,'' \emph{Sensors}, vol. 24, no. 10, Art. no. 3088, 2024.

\bibitem{balasubramaniam2022status}
A. Balasubramaniam and S. Pasricha, ``Object Detection in Autonomous Vehicles: Status and Open Challenges,'' \emph{arXiv preprint arXiv:2201.07706}, 2022.

\bibitem{benchmarkyolo}
Z. Hua, K. Aranganadin, C. C. Yeh, X. Hai, C. Y. Huang, T. C. Leung, H. Y. Hsu, Y. C. Lan, and M. C. Lin, ``A Benchmark Review of YOLO Algorithm Developments for Object Detection,'' \emph{IEEE Access}, vol. 13, pp. 123515--123545, 2025, doi: 10.1109/ACCESS.2025.3586673.

\bibitem{onestagereview}
J. Fan, T. Huo, and X. Li, ``A Review of One-Stage Detection Algorithms in Autonomous Driving,'' in \emph{Proc. 2020 4th CAA International Conference on Vehicular Control and Intelligence}, Hangzhou, China, 2020, pp. 210--214, doi: 10.1109/CVCI51460.2020.9338663.

\bibitem{yolov1toyolov11}
M. Kotthapalli, D. Ravipati, and R. Bhatia, ``YOLOv1 to YOLOv11: A Comprehensive Survey of Real-Time Object Detection Innovations and Challenges,'' \emph{arXiv preprint arXiv:2508.02067}, 2025.

\bibitem{lee2022badweather}
Y. Lee, Y. Kim, Y. Ko, J. Yu, and M. Jeon, ``Learning to Remove Bad Weather: Towards Robust Visual Perception for Self-Driving,'' \emph{IEEE Robotics and Automation Letters}, 2022.

\bibitem{liu2024largeai}
Z. Liu and H. Zhao, ``Exploration of the Application of AI Large Models in the Field of Autonomous Driving,'' in \emph{Proc. 2024 9th International Conference on Intelligent Informatics and Biomedical Sciences}, Okinawa, Japan, 2024, pp. 207--209.

\bibitem{chan2023multispectral}
H.-T. Chan, P.-T. Tsai, and C.-H. Hsia, ``Multispectral Pedestrian Detection Via Two-Stream YOLO With Complementarity Fusion For Autonomous Driving,'' in \emph{Proc. 2023 IEEE 3rd International Conference on Electronic Communications, Internet of Things and Big Data}, Taichung, Taiwan, 2023, pp. 313--316.

\bibitem{mahadshetti2024signyolo}
R. Mahadshetti, J. Kim, and T.-W. Um, ``Sign-YOLO: Traffic Sign Detection Using Attention-Based YOLOv7,'' \emph{IEEE Access}, vol. 12, pp. 132689--132700, 2024.

\bibitem{redmon2016yolo}
J. Redmon, S. Divvala, R. Girshick, and A. Farhadi, ``You Only Look Once: Unified, Real-Time Object Detection,'' in \emph{Proc. IEEE Conference on Computer Vision and Pattern Recognition}, 2016, pp. 779--788.

\bibitem{redmon2017yolo9000}
J. Redmon and A. Farhadi, ``YOLO9000: Better, Faster, Stronger,'' in \emph{Proc. IEEE Conference on Computer Vision and Pattern Recognition}, 2017, pp. 7263--7271.

\bibitem{redmon2018yolov3}
J. Redmon and A. Farhadi, ``YOLOv3: An Incremental Improvement,'' \emph{arXiv preprint arXiv:1804.02767}, 2018.

\bibitem{bochkovskiy2020yolov4}
A. Bochkovskiy, C.-Y. Wang, and H.-Y. M. Liao, ``YOLOv4: Optimal Speed and Accuracy of Object Detection,'' \emph{arXiv preprint arXiv:2004.10934}, 2020.

\bibitem{ge2021yolox}
Z. Ge, S. Liu, F. Wang, Z. Li, and J. Sun, ``YOLOX: Exceeding YOLO Series in 2021,'' \emph{arXiv preprint arXiv:2107.08430}, 2021.

\bibitem{wang2023yolov7}
C.-Y. Wang, A. Bochkovskiy, and H.-Y. M. Liao, ``YOLOv7: Trainable Bag-of-Freebies Sets New State-of-the-Art for Real-Time Object Detectors,'' in \emph{Proc. IEEE/CVF Conference on Computer Vision and Pattern Recognition}, 2023, pp. 7464--7475.

\bibitem{wang2024yolov10}
A. Wang et al., ``YOLOv10: Real-Time End-to-End Object Detection,'' \emph{arXiv preprint arXiv:2405.14458}, 2024.

\bibitem{liu2016ssd}
W. Liu et al., ``SSD: Single Shot MultiBox Detector,'' in \emph{Proc. European Conference on Computer Vision}, 2016, pp. 21--37.

\bibitem{fu2017dssd}
C.-Y. Fu, W. Liu, A. Ranga, A. Tyagi, and A. C. Berg, ``DSSD: Deconvolutional Single Shot Detector,'' \emph{arXiv preprint arXiv:1701.06659}, 2017.

\bibitem{zhang2018refinedet}
S. Zhang, L. Wen, X. Bian, Z. Lei, and S. Z. Li, ``Single-Shot Refinement Neural Network for Object Detection,'' in \emph{Proc. IEEE Conference on Computer Vision and Pattern Recognition}, 2018, pp. 4203--4212.

\bibitem{law2018cornernet}
H. Law and J. Deng, ``CornerNet: Detecting Objects as Paired Keypoints,'' in \emph{Proc. European Conference on Computer Vision}, 2018, pp. 734--750.

\bibitem{zhou2019objects}
X. Zhou, D. Wang, and P. Krahenbuhl, ``Objects as Points,'' \emph{arXiv preprint arXiv:1904.07850}, 2019.

\bibitem{tian2019fcos}
Z. Tian, C. Shen, H. Chen, and T. He, ``FCOS: Fully Convolutional One-Stage Object Detection,'' in \emph{Proc. IEEE/CVF International Conference on Computer Vision}, 2019, pp. 9627--9636.

\bibitem{lin2017focalloss}
T.-Y. Lin, P. Goyal, R. Girshick, K. He, and P. Dollar, ``Focal Loss for Dense Object Detection,'' in \emph{Proc. IEEE International Conference on Computer Vision}, 2017, pp. 2980--2988.

\bibitem{li2020gfl}
X. Li, W. Wang, L. Wu, S. Chen, X. Hu, J. Li, J. Tang, and J. Yang, ``Generalized Focal Loss: Learning Qualified and Distributed Bounding Boxes for Dense Object Detection,'' in \emph{Proc. Advances in Neural Information Processing Systems}, 2020.

\bibitem{zhang2021vfnet}
H. Zhang, Y. Wang, F. Dayoub, and N. Sunderhauf, ``VarifocalNet: An IoU-Aware Dense Object Detector,'' in \emph{Proc. IEEE/CVF Conference on Computer Vision and Pattern Recognition}, 2021, pp. 8514--8523.

\end{thebibliography}
\end{document}